%% file: root.tex
\documentclass[letterpaper, 10 pt, conference]{ieeeconf}  
\pdfoutput=1

\IEEEoverridecommandlockouts                              

\usepackage{graphicx} 
\usepackage{amsmath} 
\usepackage{amssymb}  
\usepackage{subcaption}
\usepackage{todonotes}

\newcommand{\FWName}{ {\sc DeepCrashTest}}

\title{\LARGE \bf
	\FWName: Turning Dashcam Videos into Virtual Crash Tests for Automated Driving Systems
}

\author{Sai Krishna Bashetty$^{1}$, Heni Ben Amor$^{1}$, Georgios Fainekos$^{1}$
	\thanks{$^{1}$Sai Krishna Bashetty, Heni Ben Amor and Georgios Fainekos are  with  the School of Computing, Informatics and Decision Systems Engineering, Arizona State University, 660 S. Mill Ave, Tempe, AZ 85281
		{\tt\small \{sbashett, hbenamor, fainekos\}  at asu.edu}}%
	\thanks{This research was partially funded by NSF awards 1350420, 1932068 and 1361926, and the NSF I/UCRC Center for Embedded Systems.}
}

\begin{document}

	\maketitle
	\thispagestyle{empty}
	\pagestyle{empty}

	\begin{abstract}

		The goal of this paper is to generate simulations with real-world collision scenarios for training and testing autonomous vehicles. We use numerous dashcam crash videos uploaded on the internet to extract valuable collision data and recreate the crash scenarios in a simulator. We tackle the problem of extracting 3D vehicle trajectories from videos recorded by an unknown and uncalibrated monocular camera source using a modular approach. A working architecture and demonstration videos along with the open-source implementation are provided with the paper.

	\end{abstract}

	\section{INTRODUCTION}
	\input{./introduction.tex}

	\section{RELATED WORK}
	\input{./related_work.tex}

	\section{PROBLEM DESCRIPTION}
	\input{./prob_form.tex}

	\section{TRAJECTORY EXTRACTION}
	\input{./architecture_stage1.tex}

	\section{TRAJECTORY SIMULATION AND TESTING}
	\begin{figure}[t]
		\centering
		\includegraphics[width=\linewidth]{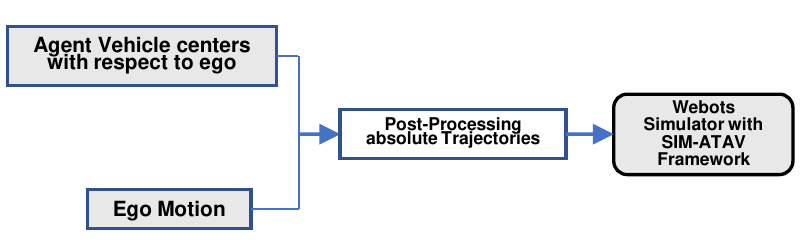}
		\caption{Vehicle Trajectory Simulation Module}
		\label{fig:architecture_stage2}
	\end{figure}
	\input{./architecture_stage2.tex}

	\section{EXPERIMENTS}
	\input{./experiments.tex}

	\section{CONCLUSIONS}
	\input{./conclusions.tex}

    \section*{ACKNOWLEDGMENT}
    We thank Dr. Pavan Turaga (ASU) and Toyota Research Institute of North America for valuable technical discussions.

	\bibliographystyle{IEEEtran}
	\bibliography{references}

%

	%
	%
	%
	%
	%
	%
	%
	%
	%
	%
	%

	%
	%
	%

\end{document}

%% file: introduction.tex
Within the last decade the field of autonomous driving has seen tremendous progress, as evidenced by the many industrial and academic entities moving at a rapid pace to win the race for full autonomy. A number of companies are deploying fleets of vehicles on public roads to collect high quality data and, in turn, train their software to ensure safe and secure transportation without human intervention. Test vehicles need to be driven approximately 11 billion miles in the real-world or a simulated environment to verify with $80\%$ confidence that they are 90 percent safer than human drivers \cite{intro_numMiles}. Among the most difficult scenarios to train and verify are unpredictable human actions (as drivers or pedestrians) that may lead to dangerous situations or accidents. 

Even though statistics \cite{axios-survey} show that in most accidents involving autonomous vehicles (AVs), the human drivers are at fault, humans are still better at handling unpredictable and potentially dangerous driving situations. In order to ensure safety and improve public trust in AV technology, it is critical to train these systems with data that deviates from nominal road behavior, e.g., encounters on the road that may lead to accidents. The current methods for collecting vehicle data primarily consists of normal driving scenarios (i.e, without abnormal driving behaviors). Extending the approach to collecting abnormal driving data, however, would be a dangerous and unsafe endeavor that would put the well-being of the test driver and his/her environment at risk. This contradiction leads to the main dilemma addressed in this paper: collecting information about hazardous road interactions is of vital importance to train and validate perception and control architectures for AVs, but is in compliance with modern ethics and safety regulations.

A common approach to circumvent this problem is by generating manually engineered simulations of driving scenarios~\cite{FremontEtAl2018arxiv}. Other approaches attempt to re-generate scenarios from existing police reports~\cite{rel_AC3R}. However, police reports show substantial variability and often lack information about critical spatial and temporal details right before the crash.  

\begin{figure}[t]
	\centering
		\includegraphics[width=\linewidth]{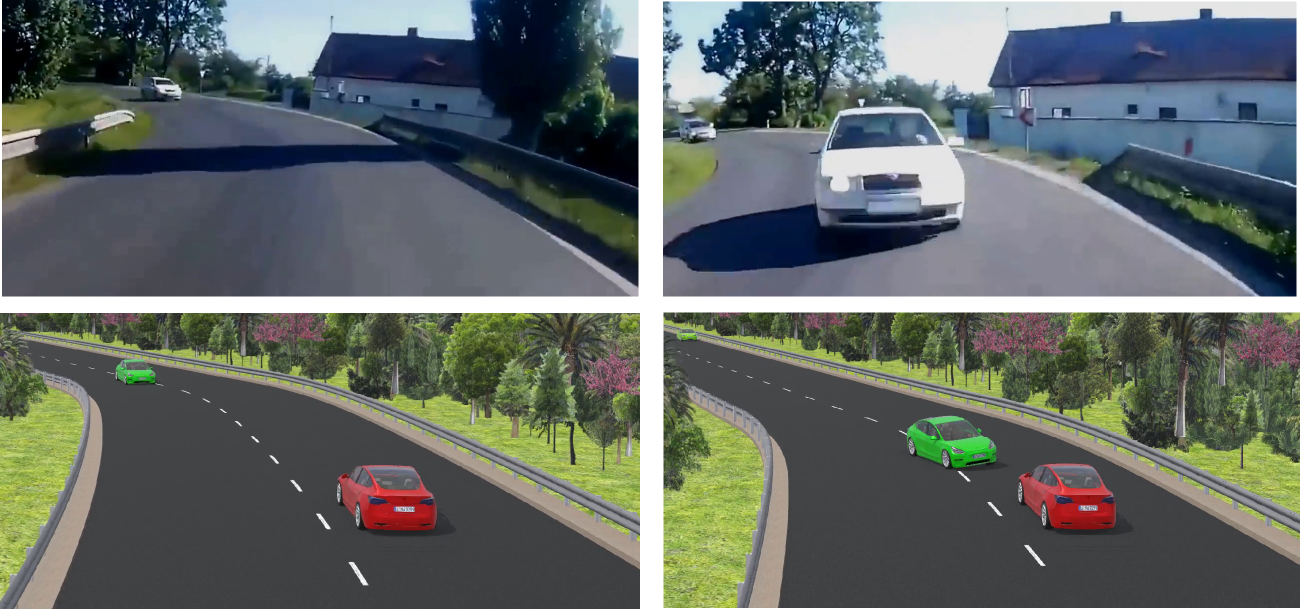} 
	\caption{Top: Two frames from a crash video at two different time instants; Bottom: the same frames in simulation}
	\label{fig:frame_to_sim}
\end{figure}

In this paper, we propose to automatically synthesize a physics based simulation of a crash by extracting relevant information from a video stream. In particular, we focus on replicating hazardous, crash-inducing behaviors found in real-world crashes. Such information can be found in abundance on the internet~\cite{intro_YouTube}, due to the proliferation of dashboard cameras. We develop a framework which can extract adversarial trajectories of vehicles from these kinds of videos and use them to automatically recreate the scene in a vehicle simulator. The simulations can be used to extract training and test data, or to study the driving behaviors leading to an accident. 
The process of extracting vehicle trajectories from data recorded with a single low quality unknown monocular sensor source is a challenging problem due to ambiguities in resolving perspective and extracting depths in monocular images. 

A modular rather than an end-to-end approach is used in this paper that utilizes multiple existing deep learning components to solve individual sub-problems.
%
The extracted trajectories are processed with custom algorithms to simulate them in the Webots simulation environment \cite{webots} using the Sim-ATAV framework \cite{sim-ATAV} (see Fig. \ref{fig:frame_to_sim}). 
Our contributions include: (i) a modular pipeline to extract 3D vehicle trajectories of the vehicles from dashcam videos, (ii) an algorithm for processing and simulating the trajectories in a vehicle simulator and (iii) \FWName~which is an add on to the Sim-ATAV testing framework \cite{assembla_link}. We also provide the simulation videos in Webots along with a demonstration of actual vehicle testing to extract safe/unsafe ego trajectories. The safe trajectories can be further analyzed to design the metrics and actions for collision avoidance systems.

%% file: related_work.tex
For detecting traffic flow from aerial video feed, implicit or explicit model-based computer vision techniques and Bayesian filters for tracking can be used \cite{rel_babinec, rel_ZuWhan}. 
Typically, such approaches require static sensors which must be initially calibrated. This assumption is orthogonal to our approach, since we want to extract detailed vehicle movement from a camera attached to the front of an (unknown) ego car. Inverse Perspective Mapping (IPM) \cite{rel_yaghoobi} can also be used to extract car positions on the road, since it generates top view images from an ego perspective. 
Unfortunately, IPM 
behaves poorly on distant objects making it hard to extract 3D trajectories.

A closely related problem in the computer vision and robotics community is Multi-body Structure From Motion (SFM) \cite{rel_sabzevari}. Multi-body SFM methods can extract vehicle trajectories from a monocular camera, but they require proper motion model approximations. Another possibility is to estimate the 6-dimensional pose~\cite{rel_poseCNN, rel_6Dpose3} of vehicles in each frame and then combine it with existing 2D tracking methods to get 3D vehicle tracks. The drawback is that 6D pose networks are designed for indoor robotic applications with a static scene and calibrated sensors.

There are works in the deep neural network literature which are closely related to the problem posed here.
Ren et al. \cite{rel_TCP} extract trajectories from static traffic cameras at an intersection using a pre-calibrated homography matrix mapping the image coordinates directly to simulator's top view.
The problem of accurate vehicle trajectory extraction can be reformulated as an end-to-end 3D object detection and tracking.
Akshay et al. \cite{rel_noblindspots} developed a real-time modular multi-object tracking system for autonomous vehicles which works with any combination of sensors. 
Markov Decision Process (MDP) are used in \cite{rel_onlineMOT} to model  a tracked object to improve long term tracking performance of agents.

Few methods like~\cite{rel_joint3D}, a network is designed by leveraging the 3D pose estimation and 2D tracking information for joint detection and tracking of vehicles using monocular video frames. 
They resort to synthetic data to overcome the deficiency in training data.
Scheidegger et al.~\cite{rel_3DMOT_pmbm} solves a similar problem by training a neural network to detect and estimate the distance to objects from a single input image. 

We use a straight-forward, yet powerful modular approach with existing architectures which only needs a monocular camera source and generalizes well to arbitrary dynamic scenes. In our approach, the problem of absolute trajectory extraction is divided into monocular 2D-object tracking and 3D-bounding box estimation for agent vehicles along with monocular visual odometry, lane tracking, and camera calibration for ego views. We believe that our modular design approach resonates well with the frameworks found in application (e.g, NVIDIA's perception pipeline \cite{rel_nvidia_drive}).


%% file: prob_form.tex
The main problem addressed in this paper can be formulated as follows: {\it Given a dashcam video captured using a monocular camera, extract the vehicle trajectories (including the ego vehicle's) and reproduce the trajectories in a vehicle simulator with a parameterization that enables further test case generation.} To solve the problem, we propose a modular architecture (a pipeline of deep neural network models) called \FWName~(see Fig. \ref{fig:architecture_stage1} for the main modules) under the following (technically necessary) assumptions:
\begin{enumerate}
	\item   Videos are short and generally under one minute in length. This is essential to avoid drift in position estimates over time because of monocular sensor effects.
	\item   The scene should be a straight or curved road as found in highways. Scenarios like intersections, have little information about the agent vehicles due to their short span of appearance with many occlusions.
	\item    It is assumed that the camera is attached to the ego car approximately at the center of the dashboard. This assumption can be relaxed with manual position adjustments of the initial simulation frame.
	\item  When the ego vehicle crashes, we only extract trajectories up to the collision time. After the crash time, the physics engine of the simulator takes over.
\end{enumerate}


%% file: architecture_stage1.tex
\begin{figure}[t]
	\centering
	\includegraphics[width=\linewidth, height=5cm]{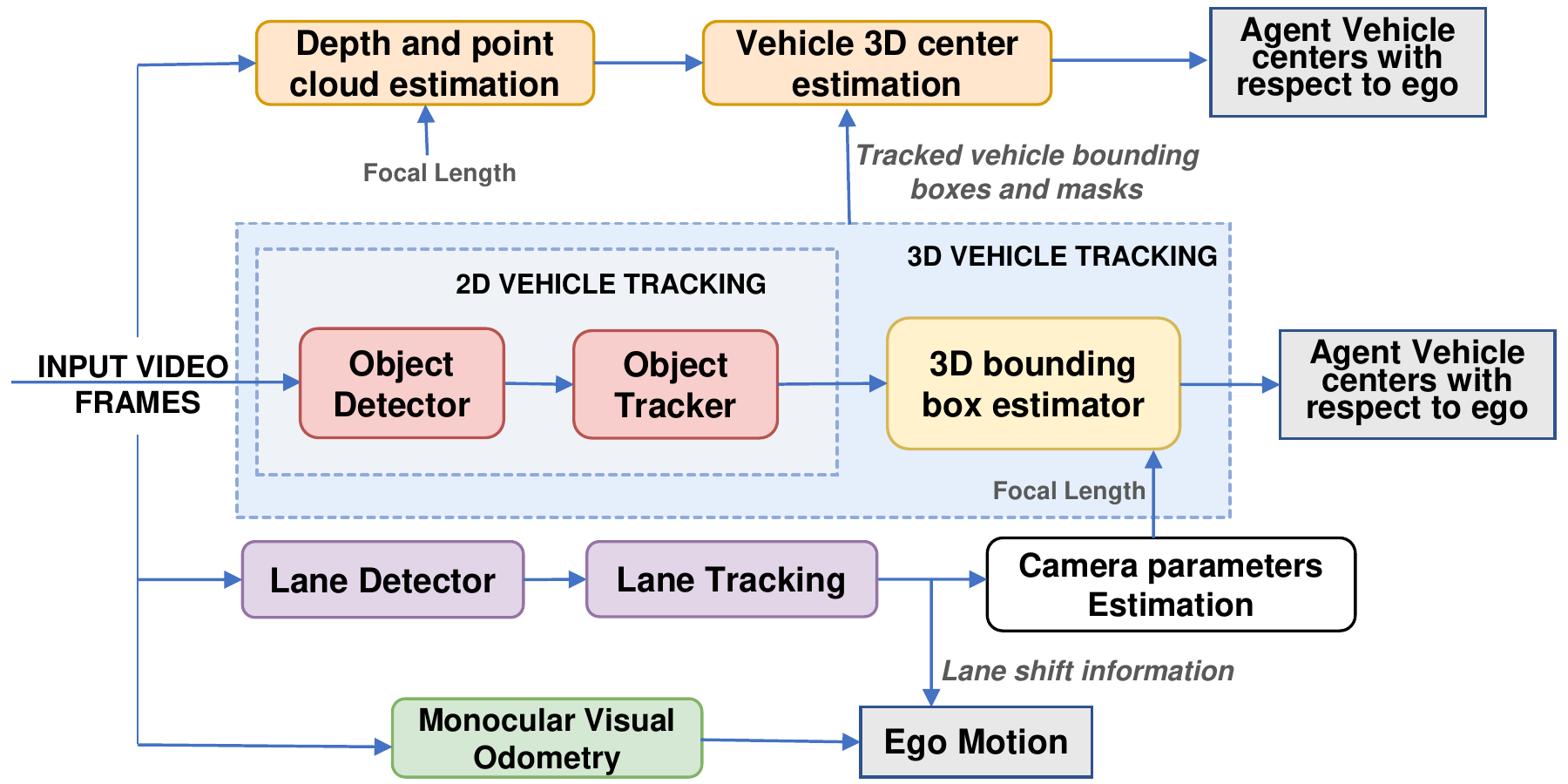}
	\caption{Vehicle Trajectory Extraction Stage}
	\label{fig:architecture_stage1}
\end{figure}

Figure \ref{fig:architecture_stage1} demonstrates our pipeline to extract 3D vehicle trajectories and individual components are explained below.

\subsection{Vehicle Detection and Tracking}

For automated vehicle tracking, we use a combination of object detector and tracker along with standard association algorithms since object trackers need annotations of initial bounding boxes for the vehicles.
There are many well known fast object detection networks such as \cite{YOLO, RCNN, Faster-RCNN, SqueezeDet} but we chose Mask-RCNN \cite{Mask-RCNN} which is an instance segmentation method which simultaneously detects objects and estimates segmented masks. Region proposal based networks like Mask-RCNN are flexible with input image dimensions which is an important criterion for our choice. Moreover, instance segmentation is necessary for an alternate approach to extract 3D bounding box centers in sec. IV-C.

For object tracking, traditional trackers \cite{obj_tracking_survey1, obj_tracking_survey2} use Kalman filters which need proper camera calibrations and fine-tuning of the motion or sensor model is generally difficult. We use the deep-learning based Re3-Tracker \cite{re3-tracker} which is robust to occlusion and has real-time capabilities.

For data association between object detector and tracker to add or remove excess trackers in a frame, we use the Hungarian algorithm \cite{hungarian} with Intersection Over Union (IOU) of bounding boxes as the matching criteria. As in \cite{git_veh_det_track}, we define threshold parameters to determine the minimum number of consecutive frames with consistent predictions for removing or adding excess trackers.

\subsection{3D bounding box Detector}
We can approximate a vehicle's trajectory (position and heading) by estimating it's sequence of 3D bounding box coordinates at each frame from the video. Using the frame rate of video, we can approximate the relative velocities of the agents with respect to ego at all time instants.
We remark that there is a high possibility of variation in the original frame rate during processing and uploading of video to internet which does not guarantee the accuracy of absolute velocity estimates. We can however safely assume that the relative trend in velocity profiles between vehicles remains unaffected. This is found to be sufficient to ensure the timing of vehicle movements and recreate the crash scenario.


Over methods like \cite{pointfusion, 3D_det_pipeline, pointcloud_3Dbox, mono3D}, we chose 3D-Deepbox network \cite{3D-deepbox} because of its simplicity and good generalization of the approach to random images. They use a VGG-Net backend with additional fully connected branches to extract the bound box dimensions along with the global orientation of the vehicles. The 3D center of the bounding box with respect to camera is further optimized by placing geometric constraints on the images.

Camera calibration is required at this stage to extract real world distances. We observed that choosing a camera intrinsic close to the dataset which the network is initially trained on (KITTI Dataset \cite{KITTI} in this case) gives a satisfactory output on quite a number of random images. We review an existing method to approximate camera parameters from lane lines in sec. IV-F.
The extracted trajectories are sent into the second stage of framework for simulation.

\subsection{Point Cloud method for 3D vehicle centers}
\begin{figure}[h]
	\centering
	\begin{subfigure}[h]{\linewidth}
		\centering
		\includegraphics[width=\linewidth]{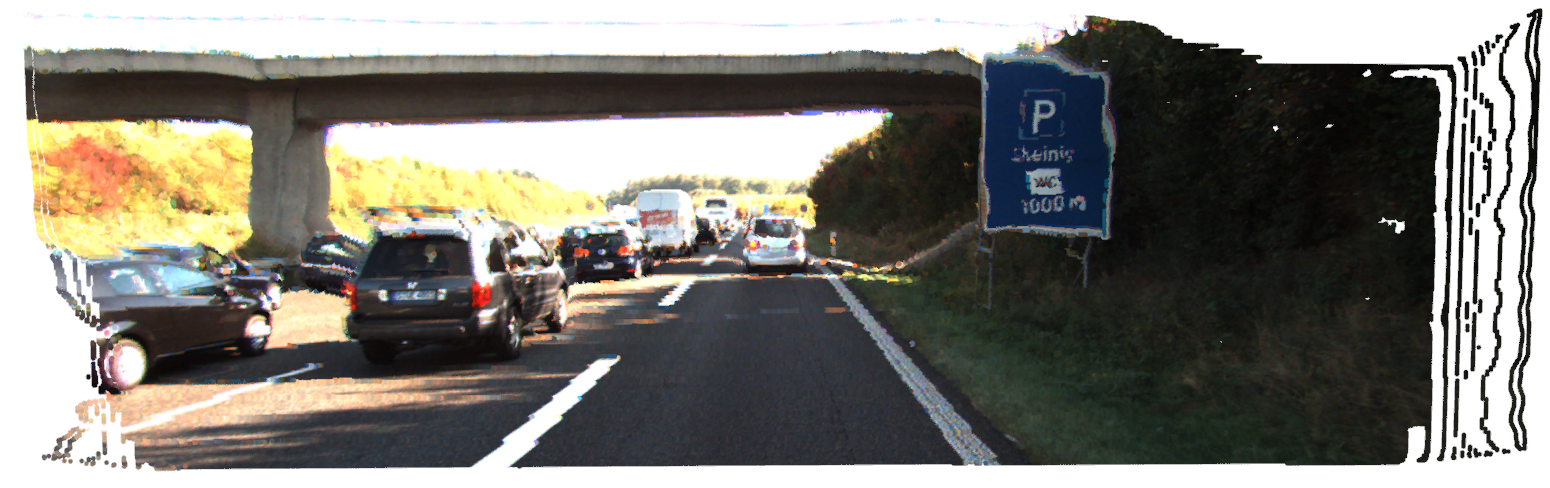}
	\end{subfigure}
	\begin{subfigure}[h]{0.48\linewidth}
		\centering
		\includegraphics[width=\linewidth]{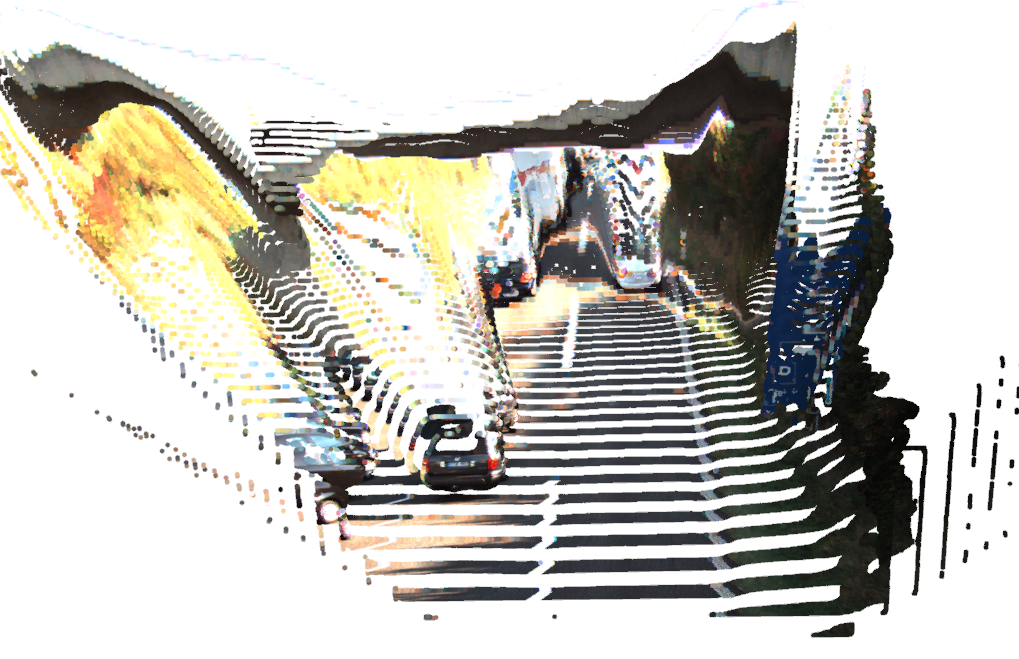}
	\end{subfigure}
	\begin{subfigure}[h]{0.48\linewidth}
		\centering
		\includegraphics[width=\linewidth]{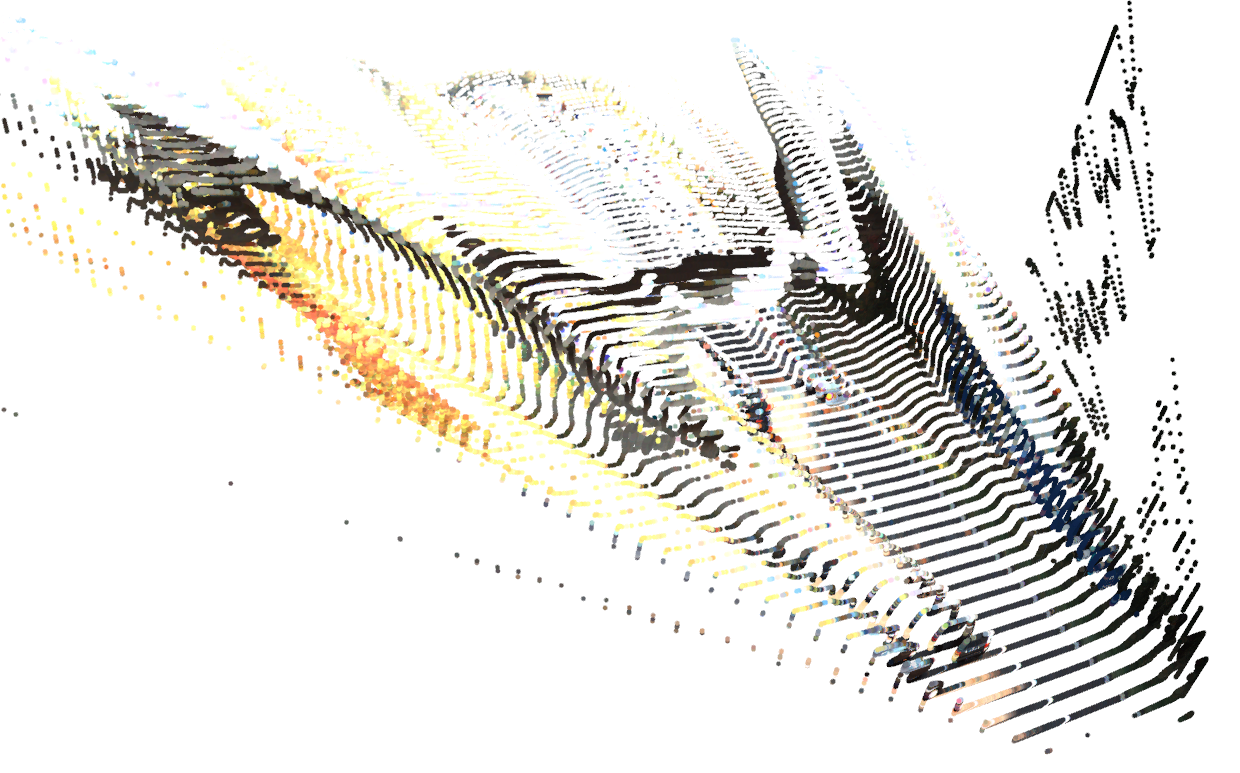}
	\end{subfigure}
	\caption{Top: Front view pseudo point cloud of an image from the KITTI dataset; Bottom: Arbitrary views of the same point cloud; The point cloud is generated from the depth images estimated using a monocular depth extraction technique \cite{monodepth}.}
	\label{fig:point_cloud}
\end{figure}

In the previous method for 3D center estimation, we have directly utilized the monocular 3D object detection networks. There is an alternative approach using depth maps of a scene for extracting only the 3D positions or specifically the distance and lateral position of the vehicle in a calibrated sensor environment with good quality images.
Given a depth map, the 3D point cloud can be generated by estimating the real world $(x,y,z)$ coordinates of individual pixels $(u,v)$ considering the depth D of scene to be in the z direction using notation and equations as in \cite{pseudolidar_from_depth} (see Fig. \ref{fig:point_cloud}):
\begin{align}
	z &= D(u,v) & x &= \dfrac{(u-c_u)\times z}{f_u} & y &= \dfrac{(v-c_v)\times z}{f_v}
\end{align}
where $(c_u, c_v)$ is the principal point and $(f_u, f_v)$ is the horizontal and vertical focal length of the camera in pixel coordinates. 
We assume the principal point to be the center of the image with equal horizontal and vertical focal length.

From the existing CNN architectures for monocular depth extraction \cite{monodepth, megadepth, depth_deep_ordinal}, we use Monodepth \cite{monodepth} because of its longer range for accurate depth prediction. Initial instance segmentations from the mask-RCNN network are used to crop the 3D point cloud corresponding to a specific vehicle. We simply find the mean of these 3D points to approximate the vehicle's position.
An important clarification to be made is that, we do not consider the 3D coordinates estimated using this approach to be the actual center of the vehicle, but just an approximate position in the scene. This is due to the fact that extracted point clouds belong only to the part of the vehicle which is exposed to the camera. However, the error in Euclidean distance between actual center and the estimated position does not exceed the dimensions of vehicle.
Hence, the path formed by this sequence of 3D points at all frames provides information about the vehicle trajectory.

Other works on similar lines are \cite{pseudolidar_from_depth, CMU_pointCloud_3D_det, pseudo_lidar++}. Their methods are much more refined and accurate since they use additional network architectures for processing. We are using a simple method which is sufficient for extracting approximate positions.
There are even end-to-end networks \cite{unsup_depth_ego, unsup_depth_ego2} for joint monocular depth extraction and ego-motion estimation but we prefer the modular approach due to it's good generalization capabilities.

\subsection{Ego Motion Estimation}
For the simulations, we need the ego vehicle's motion or camera motion to find the absolute trajectories of the agent vehicles with respect to the world frame.
In case of simple highway scenarios with the ego moving in a straight line, we can assign constant velocity and a linear path for the ego vehicle. To generalize well to arbitrary scenarios, we resorted to monocular Simultaneous Localization and Mapping (SLAM), Visual Odometry (VO), and Optical Flow based approaches to estimate the ego motion.


There are numerous works on Visual SLAM and VO and comparisons are provided by \cite{survey_slam_vo, visual_slam_survey, vo_survey}. We evaluated several methods like DSO \cite{DSO}, LSD-SLAM \cite{lsd-slam}, ORB-SLAM \cite{orbslam} and even Deep learning based approaches like Deep-VO \cite{deep-vo}. We chose the former model based approaches because of their lightweight real time capabilities.

However, we found them to introduce scale ambiguity and, also, they cannot provide real-world depth estimates which further deteriorates velocity estimates. There are a few elegant approaches built on optical flow methods which can provide good estimates with lower scale ambiguities.
Florian Raudies et al. \cite{optflow-egomotion-survey} present a survey of optical flow based ego-motion and depth estimation methods. We use the work by Andrew Jaegle et al.\cite{upennoptflow-ego}
where they introduce Expected Residual Likelihood (ERL) to remove outliers in optical flow estimates and a novel formulation of camera motion equation using lifted kernel optimization framework.

We observe that any of the above methods have stable values in the longitudinal direction of the motion of the vehicle, but they suffer from lateral drifts in the ego position. Currently, we are using lane detection and tracking methods (see next section) to localize the vehicles within lanes and prevent too much lateral drift on the road. Lane detection and tracking helps in camera calibration as well.

\subsection{Lane Detection and Tracking}
The Ego-motion estimator returns a qualitatively correct trajectory which may not be accurate enough to laterally localize the ego vehicle in the correct lane. The lane change information can even be used to completely replace the lateral ego-motion estimates in case of high drifting predictions.

Standard algorithms like the CHEVP \cite{CHEVP} or the latest ones like \cite{robust_lane_det} could be used for lane detection, but they fail when there are too many occlusions on the lane lines in high traffic scenarios and do not work well with partial lane markings. We use the deep learning based LaneNet \cite{lanenet} network which is a segmentation based approach comparatively more robust to partial or occluded lane markings.
We use a density based spatial clustering algorithm (DBSCAN) along with B-spline curve fitting with the segmented outputs of LaneNet to cluster and assign individual lane IDs.

For lane tracking, there are existing methods (e.g, \cite{lane_det_track, lane_det_track2}) involving kalman filters, but we use a simple approach where we fit straight lines through the clustered lane pixels and the lane fitting is completely in image pixel domain. We only need information about the lower part of the lane lines which is close to the ego vehicle to localize within the lanes.
The Hungarian algorithm with directed Hausdorff distance \cite {hausdorff} as a cost metric is used for association between detected and tracked lanes at each frame. 

Considering the bottom left corner of the image as the origin, we find the x-intercepts of the lane lines in pixel coordinates. Since the perspective effects or distortions in an image increase only with depth of the scene, we can safely consider that the distance between any of the x-intercepts in image space is proportional to the actual distance between the corresponding lanes.
Since, the average lane width on a specific road segment remains fairly constant, we can estimate the lateral position of the vehicle within the lanes.

\subsection{Camera Calibration}
Camera calibration is required by the 3D bounding box detection, depth extraction network and optical flow based ego-motion estimation. Due to the monocular camera setting, it is very challenging to form a closed form solution using the image features. One approach is to use  learning methods like \cite{deepfocal}\cite{focalength_esti} trained on existing datasets \cite{focalens} to estimate the focal length given an RGB image.
Fung et al. \cite{cameracalib_lane} demonstrate a more simple and light-weight approach which is the reformulation of perspective camera equations and using the lane line annotations to estimate the camera parameters.
We use this method only when the predicted lane lines are pretty accurate and in the case of errors, we use the parameters from the datasets used for training the networks. We decide this based on the visual correctness of the final 3D bounding boxes when plotted on the video frames.

%% file: architecture_stage2.tex
For all the simulations, the global (or world) origin is defined at the initial position of the ego. The absolute positions of all the vehicles throughout the simulation are calculated with respect to this origin.

\subsection{Trajectory smoothing}
We use a Savitzky-Golay filter followed by a spline smoothing for the raw trajectories. Two levels of spline smoothing is used, first is local windowed with high smoothness factor and, then, a global fitting with low smoothing factor. This ensures an overall smooth curve while still preserving the sudden abnormal vehicle movements that lead to collisions. Moreover, splines provide the needed trajectory parameterization to use test generation frameworks such as sim-ATAV \cite{sim-ATAV}.
The extracted lateral localization from lane predictions is added to the corresponding ego position at each frame in this stage.

\subsection{Extrapolating Vehicle Trajectories}
The agent vehicle trajectories can be categorized based on their heading direction and the time of appearance or disappearance from the scene. Agent vehicles can either move in same or opposite direction of ego i.e, ongoing or oncoming, respectively.

The ongoing vehicles can be classified into three types:
\\
\textbf{\textit{D0T1}}: Agent vehicle is in the scene since the first frame.
\\
\textbf{\textit{D0T2}}: Agent vehicle enters later into the scene from behind i.e, agent overtaking the ego.
\\
\textbf{\textit{D0T3}}: Agent vehicle enters later from front in the scene i.e, a far off vehicle slowing down or pulling over or stopped by the roadside. This can even be considered as a case where the ego overtakes an agent vehicle which is initially far ahead from the ego.

The oncoming vehicles can be classified into four types:
\\
\textbf{\textit{ D1T1}}: Agent vehicle already in scene since the first frame and the ego vehicle eventually passes it before the completion of the video sequence.
\\
\textbf{\textit{D1T2}}: Agent vehicle is in scene since the first frame and the ego vehicle {\it doesnot} pass it through the entire video sequence. This may happen when the agent vehicle collides with ego or the video clip ends before the agent reaches the ego.
\\
\textbf{\textit{D1T3}}: Agent vehicle enters later into the scene and the ego passes it.
\\
\textbf{\textit{D1T4}}: Agent vehicle enters later into the scene and the ego {\it doesnot} pass it.

This kind of classification is necessary for appropriate trajectory processing and is made purely based on our observations of the dashcam crash videos.
\newline
\textit{Processing Ongoing vehicle trajectories:} 
\textit{D0T1} paths can be directly used without any processing, but the \textit{D0T2} paths need extrapolation of path and velocity profiles. Delay needs to be added to \textit{D0T3} vehicles such that the frame and position at which they appear in the scene with respect to ego is properly synchronized with the original video. 
For \textit{D0T2} vehicles, until the frame where they appear in the scene, their initial positions and velocity profile are extrapolated such that the agent mimics the ego motion. Sufficient distance is maintained such that it does not collide with ego or does not appear in the scene.
\textit{D0T3} vehicles are initially far ahead of ego and appear much later in the video. To recreate this, we initialize agents at their initial position and then introduce some delay in the controller calculated based on frame of appearance and frame rate of video.
In all the ongoing vehicle trajectories, if the ego overtakes other vehicles and the agent leaves scene before end of video, we extrapolate these trajectories until the end of the simulation following the road or lane at a velocity less than the ego.
\newline
\textit{Processing oncoming vehicle trajectories:}
The \textit{D1T1} and \textit{D1T3} vehicles are passed by the ego in the video and we do not have information once they move out of the scene. We extrapolate these trajectories such that they follow the road with a constant velocity after they pass the ego.
\textit{D1T3} and \textit{D1T4} enter later into the scene and similar to \textit{D0T3} vehicles, we generate initial start delays.

\subsection{Generating Road Waypoints}
Generating proper road structure is important for recreating plausible simulation. In most of the cases, since the crash happens on an highway, we can simply generate a straight road structure. However, it is also important to reproduce curved roads as observed in one of our video demonstrations which involves an oncoming collision on a sharp turning.
We use the ego vehicle's qualitative trajectory as reference to generate the road structure and perform spline smoothing with a high smoothness factor. For oncoming collisions, the ego trajectory does not have any information about the road structure beyond the point of collision. However, it is necessary to extend the road beyond the collision point. One such example can be observed in the head on collision simulation in our video submission. Similar to the road structure generated using ego vehicle paths, we use the oncoming agent paths to extend the road by using spline interpolation to generate a continuous road structure. The entire process is automated in the framework. 
This approach does not guarantee accurate road estimation but is found to generate qualitatively correct simulations.

\subsection{Generating Trajectories in Webots}
Typically, in the available video clips, the vehicles move with non zero velocity from the initial frame. Since starting vehicles with non-zero velocities in Webots simulations does not result in convincing natural movements, we need the vehicles from zero initial velocity until all the vehicles reach the intended velocities and relative positions at the right time.

The time taken or distance traveled in a straight path to reach that target velocity under constant acceleration can be directly calculated from kinematic equations. We will refer to this as step-back time and step-back distance, respectively. Since the target velocities of all the vehicles where they first appear in the scene are different, each vehicle needs varying step-back times and distances. Additionally, we need to synchronize the relative positions of all the vehicles with those estimated in the initial frame of the video by the time they reach their respective target velocities.

To ensure this, we first determine the maximum of all step-back times $t^{s}_{max} = \max{(t^{s}_{1}, t^{s}_{2},..., t^{s}_{n})}$, compute the total step-back distance $D^{s}_{i}$ for each vehicle $i$ with step-back time $t^{s}_{i}$ as $D^{s}_{i}$ = $d^{s}_{i} + (t^{s}_{max} - t^{s}_{i})*v^{t}_{i}$, where $d^{s}_{i}$ is the step-back distance for vehicle $i$ to reach its target velocity $v^{t}_{i}$, estimated for its initial frame of appearance and $n$ is the number of vehicles of interest to be simulated. We use the calculated distances as the length for extending the initial segment of the paths. The vehicles follow a straight path until they merge with their actual velocity and path profiles in the video. The part of the simulation which represents or recreates the actual video starts after time $t^{s}_{max}$.


Additionally, we provide visualization and editing tools for minor modifications to the lateral or longitudinal displacements of individual trajectory waypoints either to easily change the initial conditions or to ensure that vehicles do not overlap during initialization. We reiterate that since our target application is testing in a virtual environment which is going to be further modified, we do not need a quantitatively accurate reproduction of the real driving scenario. In other words, a visual inspection for qualitative correctness is sufficient for our application.

%% file: experiments.tex
\begin{table}[t]
    \caption{The table provides the modified (refer Sec. VI-B) KITTI tracking benchmark results on three KITTI \cite{KITTI} tracking sequences 3, 8, 10}
    \label{table:kitti_evaluations}

    \begin{tabular}{c | c | c | c | c | c}
    {\bf Sequences} & {\bf MOTA} & {\bf MOTP} & {\bf MODA} & {\bf MODP} & {\bf recall}\\ \hline
    SEQ 3 & 58.08 \% & 80.35 \% & 59.28 \% & 83.33 \% & 78.17 \% \\
    SEQ 8 & 67.36 \% & 79.55 \% & 67.36 \% & 84.54 \% & 72.08 \% \\
    SEQ 10 & 77.75 \% & 85.04 \% & 78.44 \% & 86.96 \% & 83.01 \% \\
    \end{tabular}
    
    \begin{tabular}{c | c | c | c | c | c | c}
    {\bf Sequences} & {\bf precision} & {\bf F1} & {\bf TP} & {\bf FP} & {\bf FN} & {\bf FAR}\\ \hline
    SEQ 3 & 81.03 \% & 79.75 \% & 265 & 62 & 74 & 42.75 \% \\
    SEQ 8 & 94.35 \% & 81.73 \% & 736 & 44 & 285 & 11.25 \% \\
    SEQ 10 & 94.90 \% & 88.56 \% & 484 & 26 & 99 & 8.81 \% \\
    \end{tabular}

    \begin{tabular}{c | c | c | c | c | c}
    {\bf Sequences} & {\bf objects} & {\bf trajectories} & {\bf MT} & {\bf PT} & {\bf ML}\\ \hline
    SEQ 3 & 355 & 10 & 37.5 \% & 62.5 \% & 0.00 \% \\
    SEQ 8 & 796 & 21 & 47.61 \% & 38.09 \% & 14.28 \% \\
    SEQ 10 & 510 & 14 & 23.07 \% & 76.92 \% & 0.00 \% \\
    \end{tabular}
\end{table}

\subsection{Qualitative Evaluation}
In this section, we describe our procedure for a qualitative assessment of our framework through a typical usage of the extracted adversarial trajectories in a  test environment for a collision avoidance system. 
We use a naive Automatic Emergency Braking (AEB) controller from the Sim-ATAV~\cite{sim-ATAV} framework. 
The AEB in sim-ATAV uses a sensor fusion module to perceive the environment by utilizing the sensors attached to the ego vehicle. In our tests, we activate the GPS/IMU, monocular dashcam and a single Radar sensor attached to the front of the ego vehicle to localize the agents in the scene.

\begin{figure}[t]
	\centering
	\begin{minipage}{\linewidth}
		\begin{subfigure}[h]{0.34\linewidth}
			\centering
			\includegraphics[width=\linewidth, height=2.8cm]{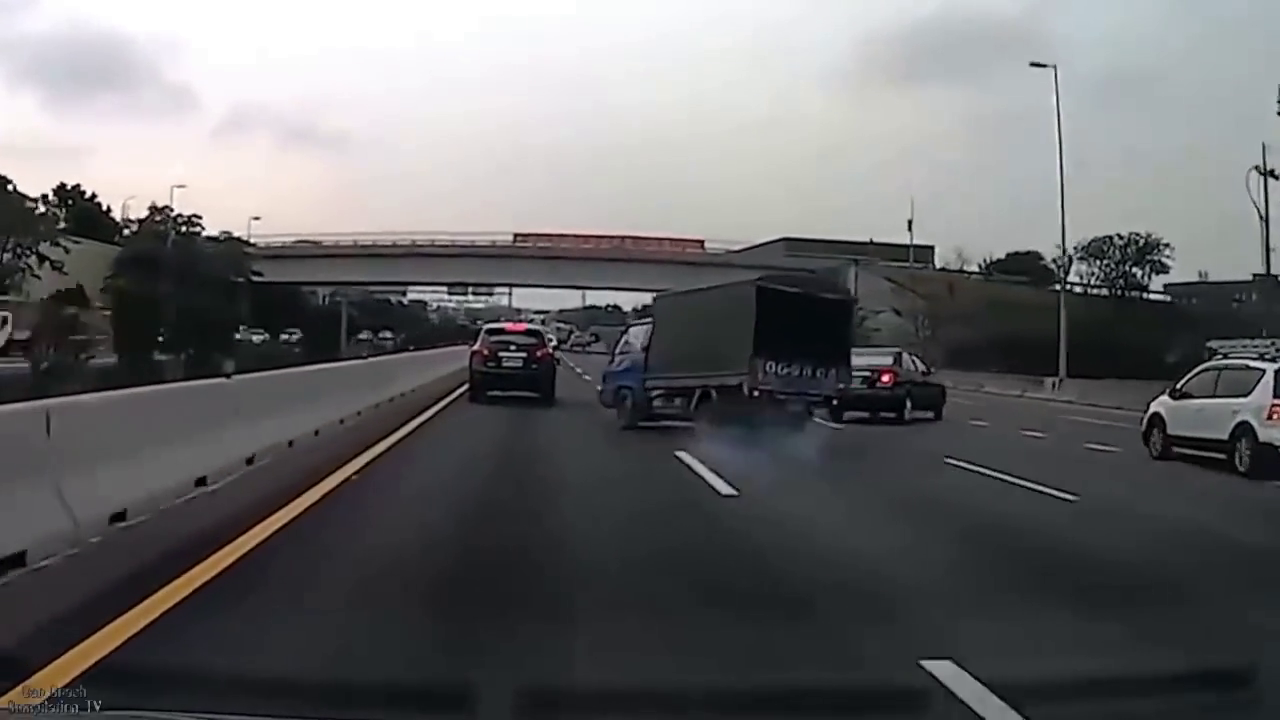} 
		\end{subfigure}
		\begin{subfigure}[h]{0.32\linewidth}
			\centering
			\includegraphics[width=\linewidth, height=2.8cm]{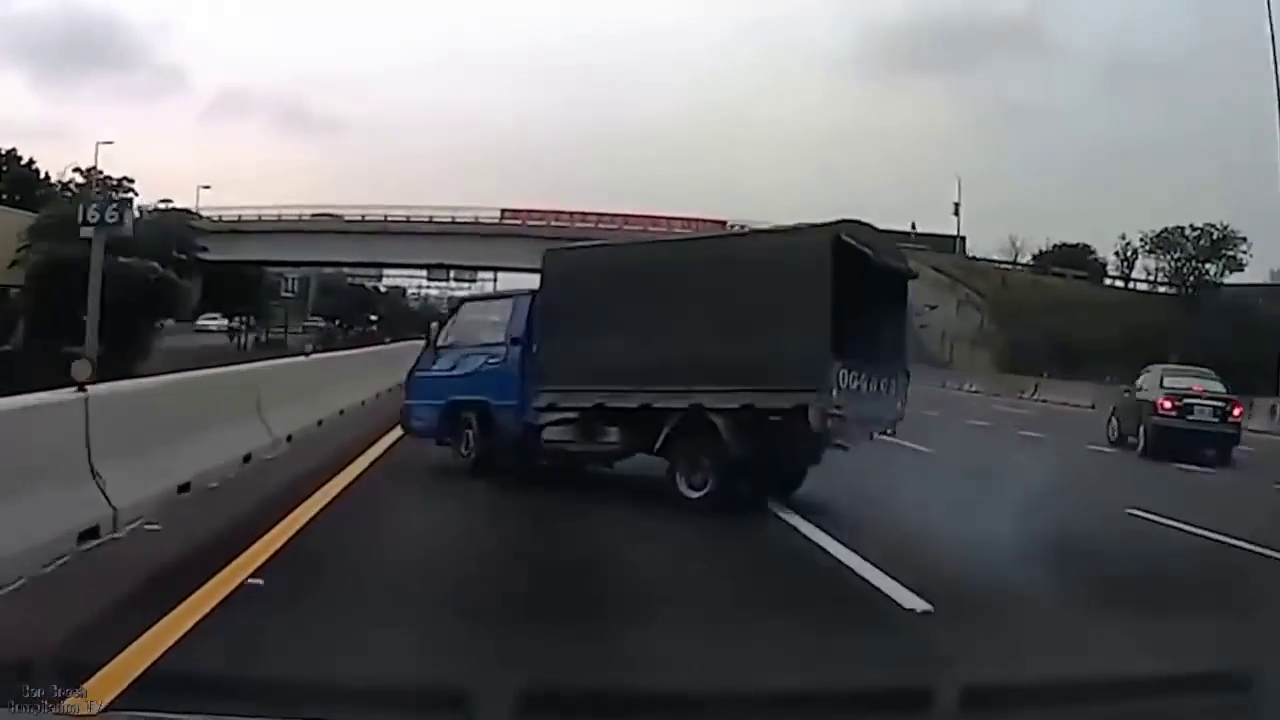}  
		\end{subfigure}
		\begin{subfigure}[h]{0.32\linewidth}
			\centering
			\includegraphics[width=\linewidth, height=2.8cm]{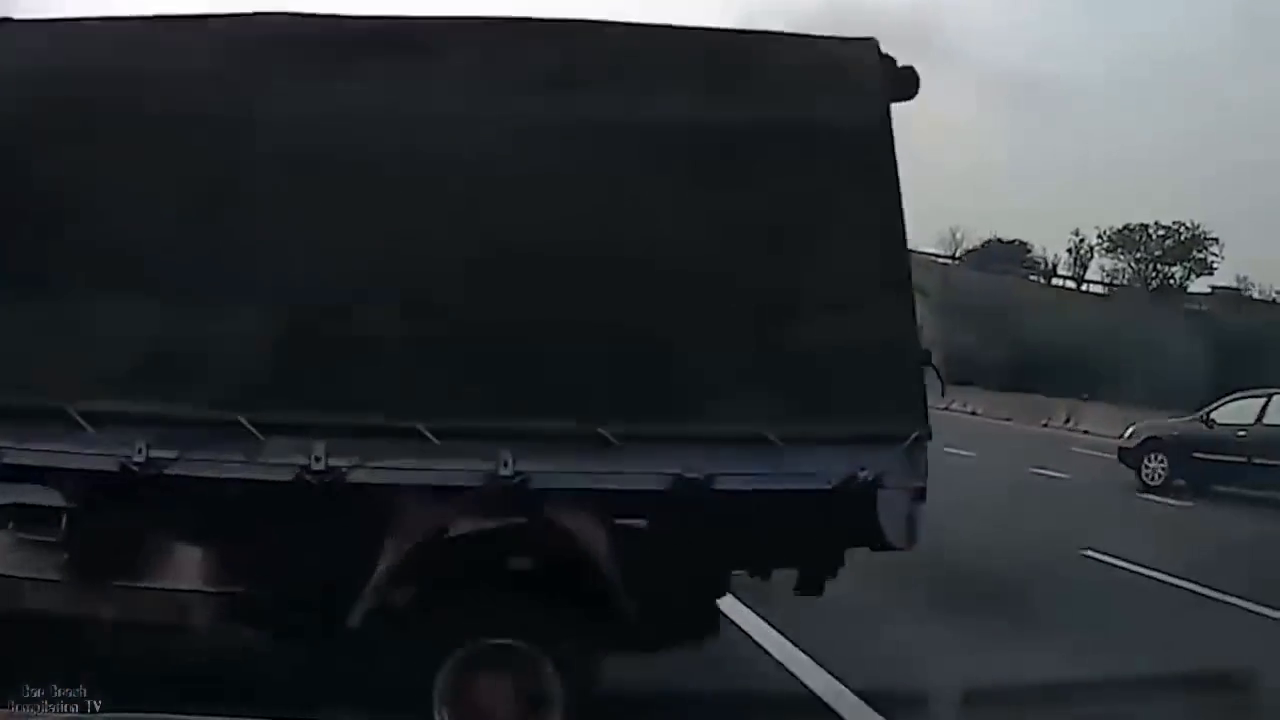}
		
		\end{subfigure}
		\subcaption{}
		\label{fig:eval_original}
	\end{minipage}
	\begin{minipage}{\linewidth}
		\begin{subfigure}[h]{0.34\linewidth}
			\centering
			\includegraphics[width=\linewidth, height=2.8cm]{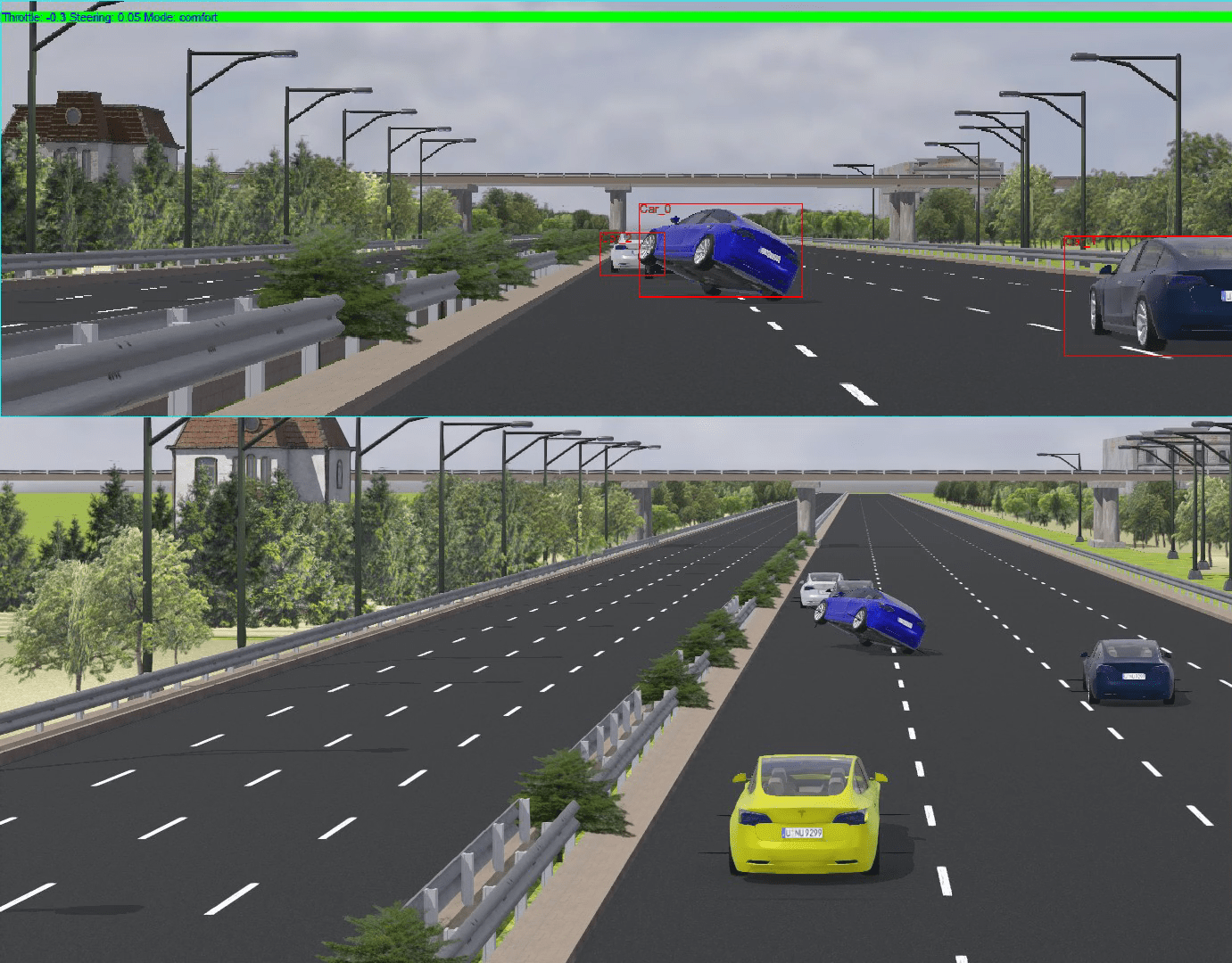}
		\end{subfigure}
		\begin{subfigure}[h]{0.32\linewidth}
			\centering
			\includegraphics[width=\linewidth, height=2.8cm]{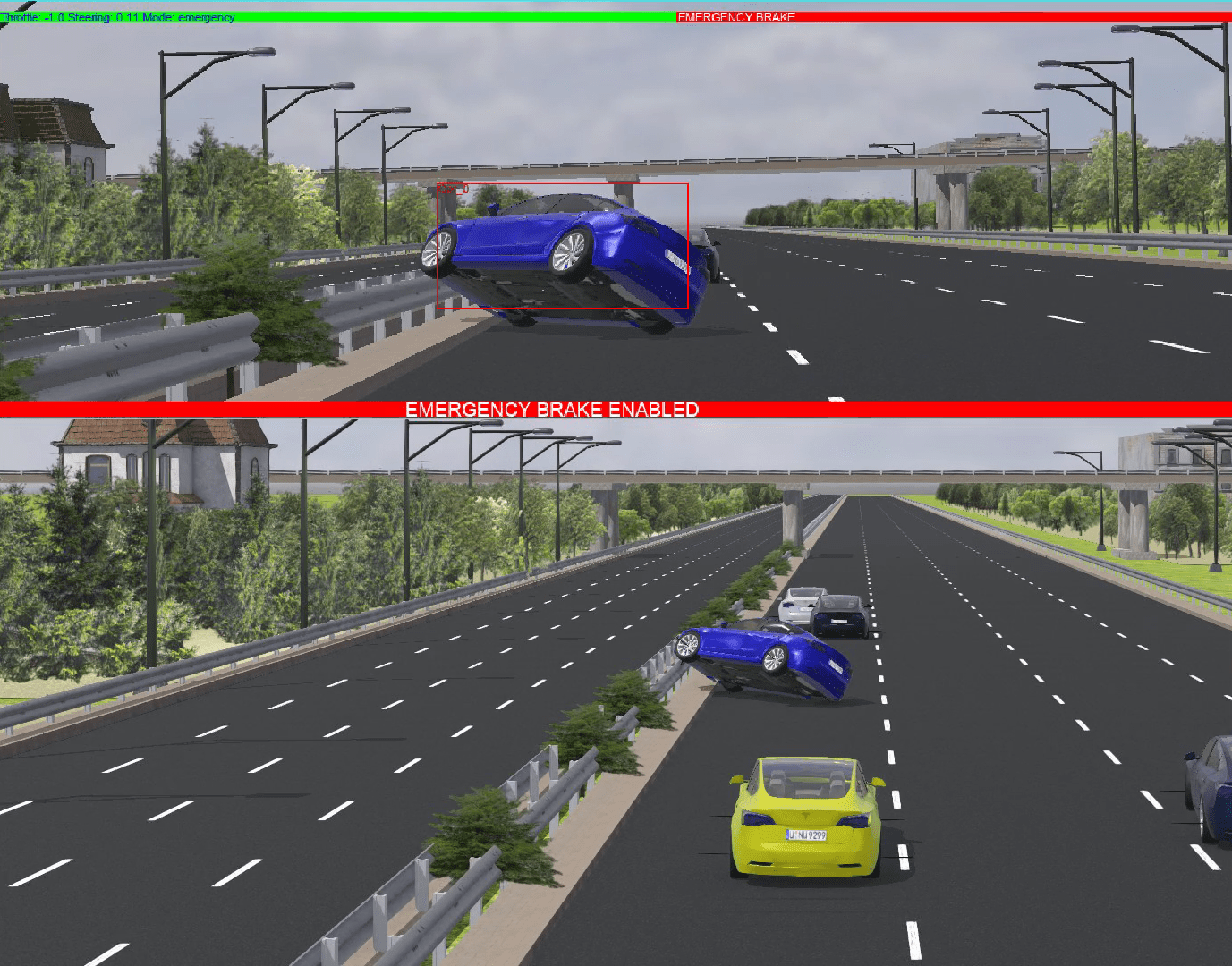}
		\end{subfigure}
		\begin{subfigure}[h]{0.32\linewidth}
			\centering
			\includegraphics[width=\linewidth, height=2.8cm]{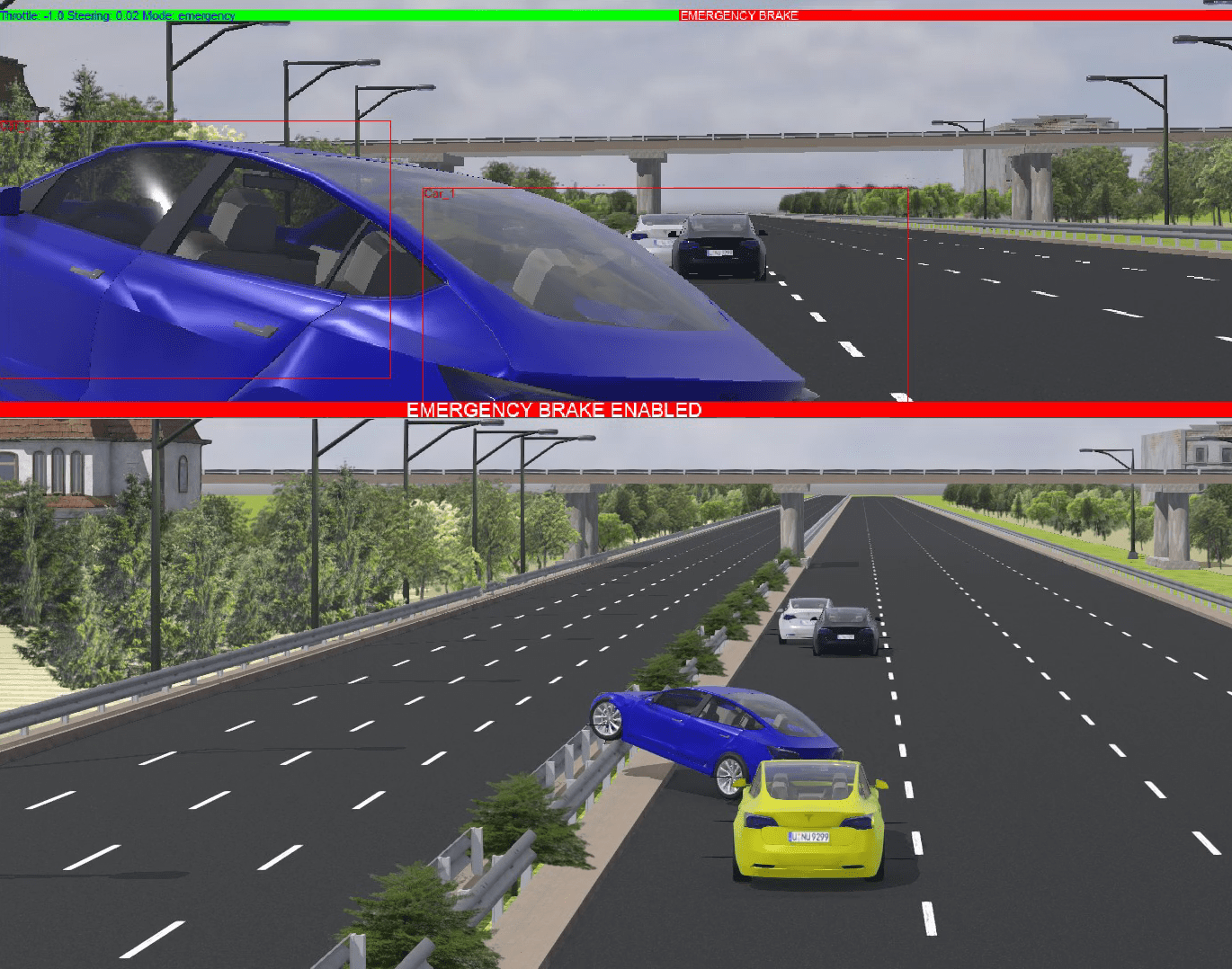}          
		\end{subfigure}
			\subcaption{}
		\label{fig:eval_sim_unsafe}
	\end{minipage}
	\begin{minipage}{\linewidth}
		\begin{subfigure}[h]{0.34\linewidth}
			\centering
			\includegraphics[width=\linewidth, height=2.8cm]{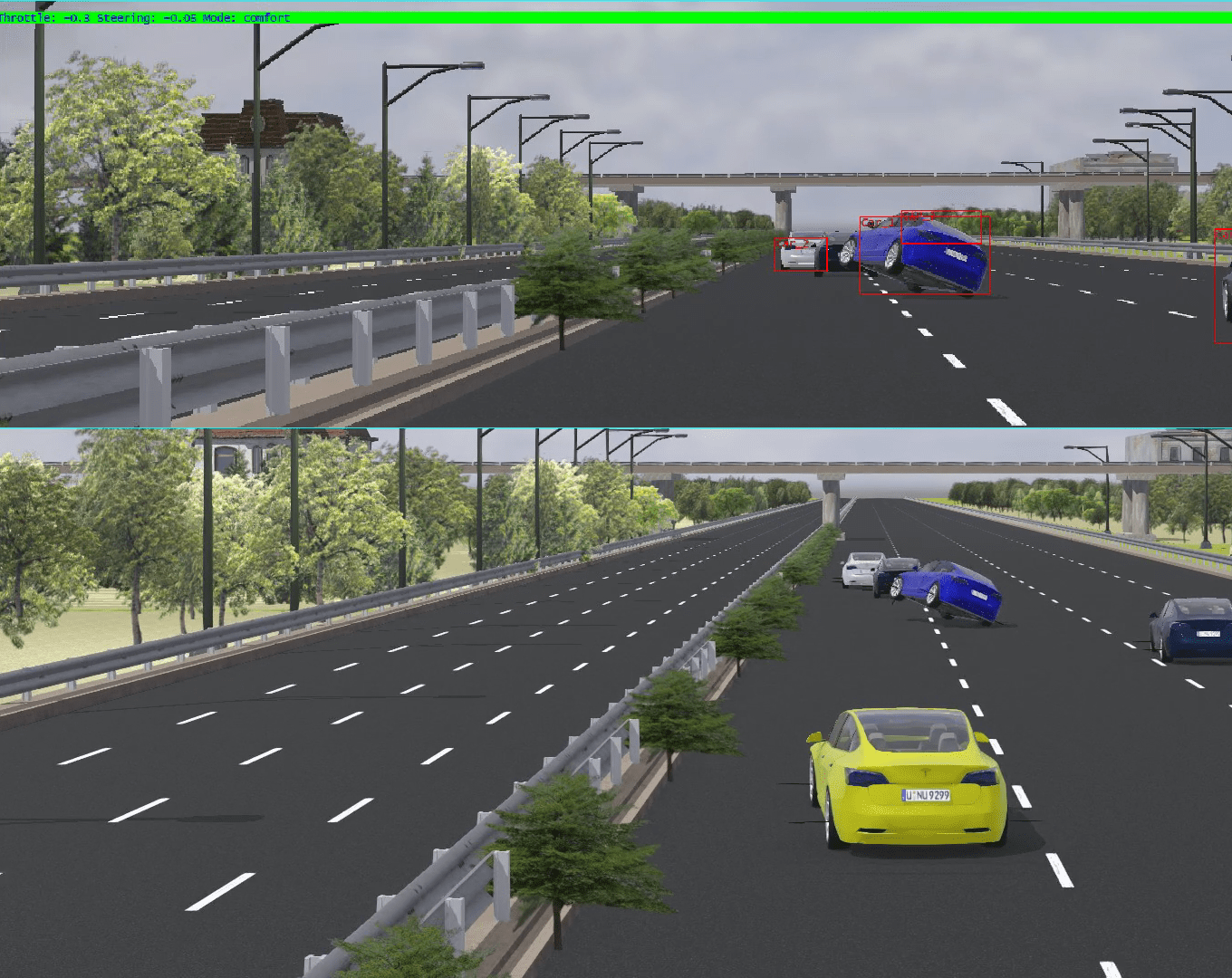}
		\end{subfigure}
		\begin{subfigure}[h]{0.32\linewidth}
			\centering
			\includegraphics[width=\linewidth, height=2.8cm]{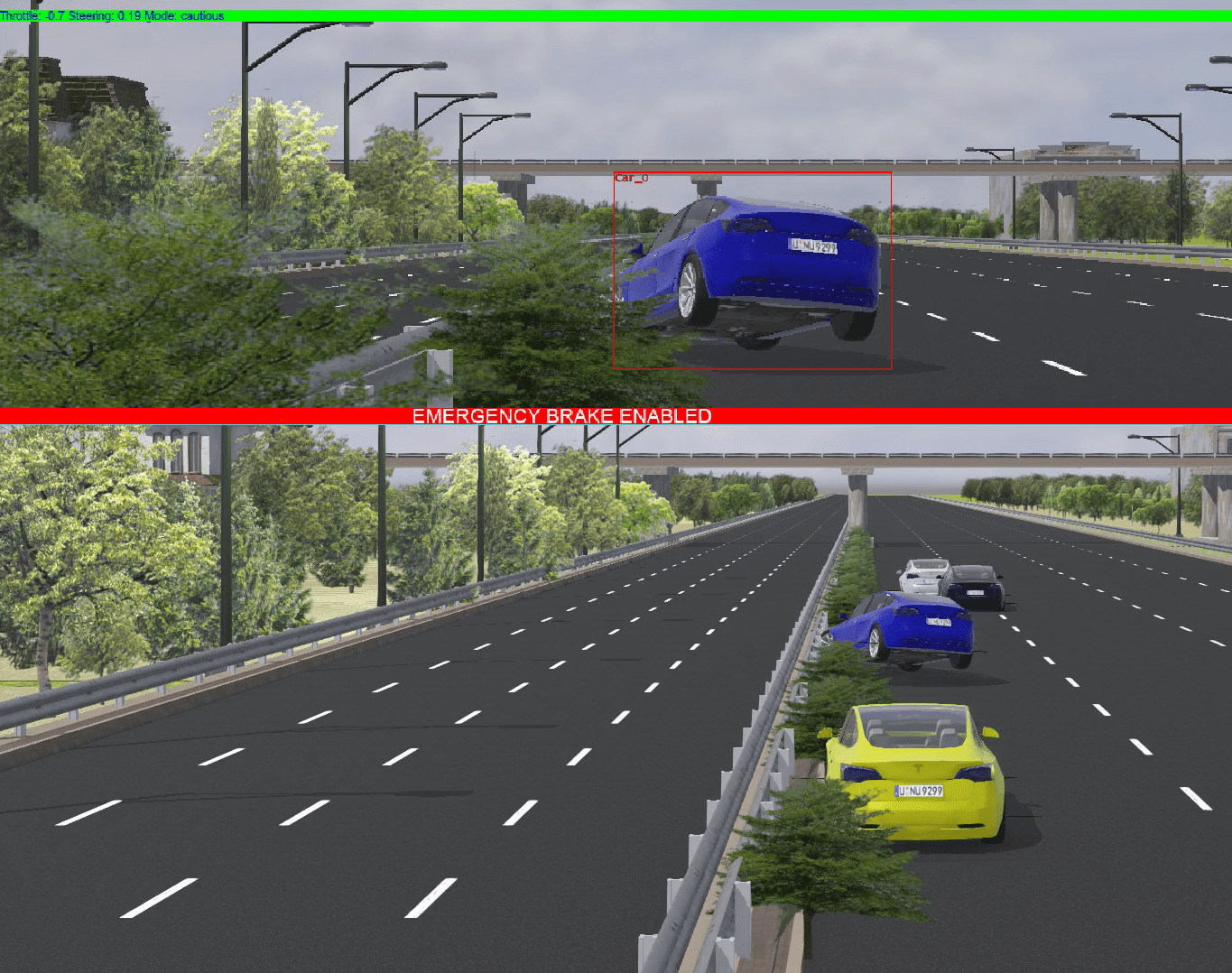}
		\end{subfigure}
		\begin{subfigure}[h]{0.32\linewidth}
			\centering
			\includegraphics[width=\linewidth, height=2.8cm]{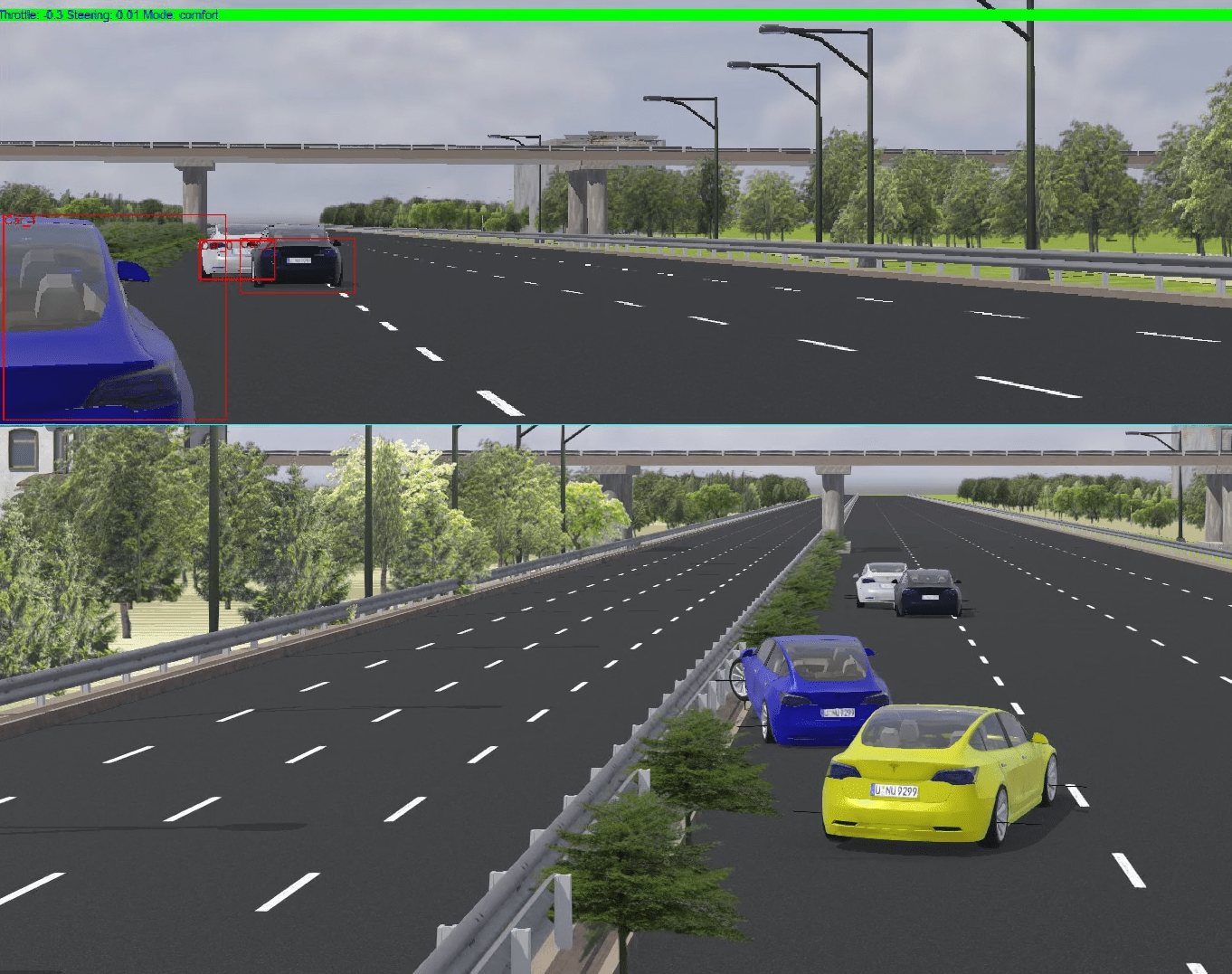}        
		\end{subfigure}
			\subcaption{}
		\label{fig:eval_sim_safe}
	\end{minipage}	
	\caption{ (a): Three frames from a crash video at different time instants; (b): The corresponding frames in simulation with original trajectories; (c): The same frames in simulation with safe ego behavior without collision which is generated by randomly searching various initial conditions.}
	\label{fig:evaluations}
	
\end{figure}

Figure \ref{fig:eval_original} shows three time ordered images of an original crash video. Fig.~\ref{fig:eval_sim_unsafe} includes the corresponding frames in a simulation generated from the trajectories extracted using \FWName~on the original video. The second and third images in Fig. \ref{fig:eval_sim_unsafe} show that though the AEB activates before the collision, it collides with the agent. This is because, the velocity of ego vehicle is very high (around 90 Kmph) and the distance to obstacle is very low compared to the required braking distance at such high velocities. At these velocities only way to completely avoid a collision is to steer away from the obstacle. To extract safe maneuvers of the ego vehicle which avoids a collision, we generated 128 simulations. For these simulations, we first manually sampled 32 initial positions within a box of 4x8(m) around the original position of ego. For each of these initial positions, we added small 2D guassian noise to the spline control points of the original ego vehicle trajectory to 4 random trajectories. Eight of the 128 simulations did not have a collision. We picked four meaningful ego motions and Fig. \ref{fig:eval_sim_safe} includes frames sampled from one of the safe simulation. As can be seen from the third image of \ref{fig:eval_sim_safe}, before collision, the ego steers away from the agent thus avoiding collision.

\subsection{Quantitative Evaluation}
Even though, our application focuses on qualitative performance, we provide quantitative evaluations for the trajectory extraction module (see fig. \ref{fig:architecture_stage1}). This is not straightforward since, \FWName~deals with an unknown dashcam video without ground truth, however, we provide evaluations by testing the framework with the KITTI tracking dataset \cite{KITTI}. The KITTI benchmark provides tracking evaluation where the object positions are all relative to the ego vehicle but we want to evaluate the absolute trajectory positions by including the ego motion. For this, we added the ground truth odometry information to the ground truth tracking data to evaluate our framework. Table \ref{table:kitti_evaluations} provides the tracking evaluation results on vehicles using such a modified version for three tracking sequences from the dataset (sequences 3, 8, 10). The actual performance can be slightly lower because of errors in camera calibrations.

%% file: conclusions.tex
We developed a framework called \FWName~for extracting 3D vehicle trajectories from dashcam videos uploaded to the internet.
Currently, \FWName~is an add-on to the test generation framework Sim-ATAV \cite{sim-ATAV} within which we recreate the extracted scenarios. 
 \FWName~can be used for testing collision avoidance systems, or more generally autonomous vehicles. 
 We demonstrated a specific use-case for extracting safe/unsafe vehicle trajectories within Sim-ATAV, but \FWName~could also be used as an add-on to a range of simulation-based automated test generation tools for AV, e.g., \cite{FremontEtAl2018arxiv,Abbas17Cyphy,MajumdarEtAl2019paracosm}. 
 